\documentclass[journal]{latex-templates/IEEE-Transactions-LaTeX2e/IEEEtran}

\usepackage{iftex}
\ifXeTeX
  \usepackage{fontspec}
\else
  \PackageError{OpenLoopEvolve}{The English version must be compiled with XeLaTeX}{}
\fi

\usepackage{amsmath,amssymb}
\usepackage{booktabs}
\usepackage{graphicx}
\usepackage{multirow}
\usepackage{placeins}
\usepackage{algpseudocode}
\usepackage{xcolor}
\usepackage{url}
\usepackage{cite}
\usepackage[hidelinks]{hyperref}

\algrenewcommand\algorithmicrequire{\textbf{Input:}}
\algrenewcommand\algorithmicensure{\textbf{Output:}}
\newcommand{\AlgExplain}[1]{%
  \Statex \hspace{\algorithmicindent}%
  {\scriptsize\color{black!62}$\triangleright$\enspace #1}%
}

\newcounter{olealgorithm}
\newenvironment{olealgorithm}[1]{%
  \refstepcounter{olealgorithm}%
  \par\smallskip\noindent
  \begin{minipage}{\columnwidth}
  \footnotesize
  \hrule\vspace{2pt}
  \noindent\textbf{Algorithm~\theolealgorithm}\quad #1\par
  \vspace{2pt}\hrule\vspace{2pt}
}{%
  \vspace{2pt}\hrule
  \end{minipage}
  \par\smallskip
}

\graphicspath{{figures/}}

\begin{document}

\title{OpenLoopEvolve: A Verifiable Self-Evolution Framework for Loop Policies in Long-Horizon Complex Tasks}

\author{Siqi Wang\textsuperscript{*}, Xinlin Li\textsuperscript{*}, Zhenglin Li, and Li Li%
\thanks{\textsuperscript{*}Siqi Wang and Xinlin Li contributed equally to this work and share first authorship.}%
\thanks{Siqi Wang and Li Li are with the Department of Automation, Tsinghua University (e-mail: wang-sq24@mails.tsinghua.edu.cn; li-li@mail.tsinghua.edu.cn).}%
\thanks{Xinlin Li is with the Department of Industrial Engineering, Tsinghua University (e-mail: lix122@mails.tsinghua.edu.cn).}%
\thanks{Zhenglin Li is with Shenzhen International Graduate School, Tsinghua University (e-mail: li-zl23@tsinghua.org.cn).}%
\thanks{Corresponding author: Li Li.}}

\hypersetup{pdfauthor={Siqi Wang, Xinlin Li, Zhenglin Li, and Li Li}}

\maketitle

\begin{abstract}
Long-horizon complex tasks require agents to repeatedly observe states, formulate plans, invoke tools, verify results, and recover from failures in continuously changing environments. However, such control experience often remains confined to a single context or a fixed prompt, and is difficult to accumulate and reuse across historical traces. This paper presents OpenLoopEvolve (OLE), a self-evolution framework centered on the Loop Policy. OLE represents an agent's observation, planning, memory, action, verification, recovery, stopping, and budget control as portable policy assets with versions and lineages, and provides online and offline evolution modes that can be selected according to practical needs: the online mode triggers candidate generation based on feedback from continuous operation, whereas the offline mode searches for candidate policies from archived traces and failure evidence. Both modes share an evolution mechanism consisting of autonomous proposals by a large language model, Champion--Challenger paired evaluation, and robust release. Policies released online are activated at a subsequent task boundary, monitored using subsequent feedback, and rolled back to their parent versions when degradation conditions are met. On the simulated business benchmark YC-Bench, both modes improve aggregate task performance, task success rate, and risk metrics relative to a fixed initial Loop Policy. The results indicate that treating the Loop Policy as a governable asset can support the accumulation, comparison, release, and reuse of control experience and improve agent performance on long-horizon complex tasks.
\end{abstract}


\section{Introduction}

\begin{figure*}[!t]
  \centering
  \includegraphics[width=0.93\textwidth]{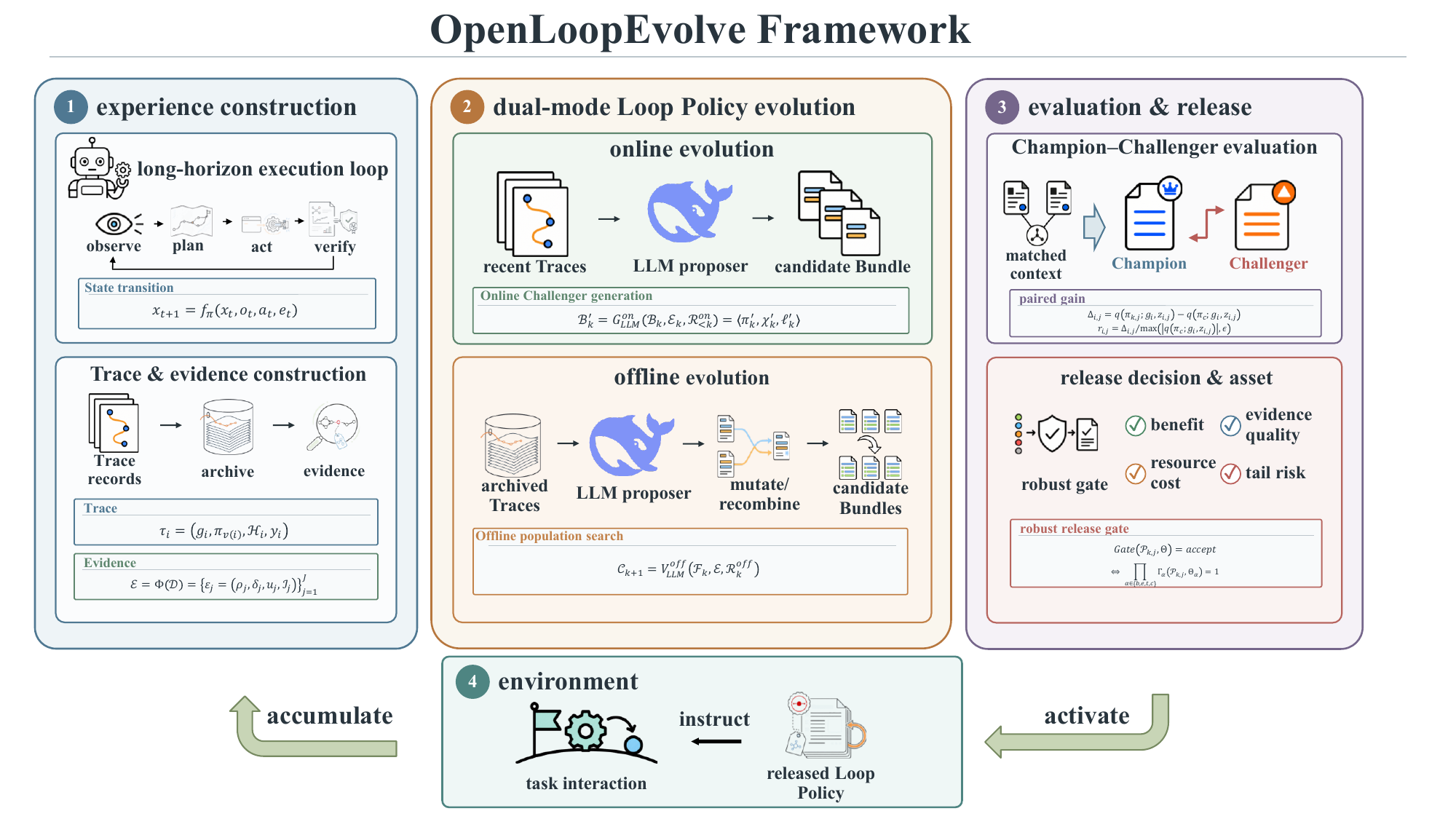}
  \caption{Overall OpenLoopEvolve framework. A released Loop Policy is activated at a subsequent task boundary, enters the host environment, and controls interactions in long-horizon tasks, producing attributable traces. The traces are converted into evolution evidence under a shared evidence contract and enter online or offline Loop Policy evolution. Candidates are evaluated against the Champion under identical conditions and, after passing the robust gate, are released as versioned policy assets and activated at a task boundary.}
  \label{fig:openloopevolve-framework}
\end{figure*}

Large language model agents are evolving from tools that generate one-shot answers into task executors capable of sustained action in external environments. Across a sequence of decisions in a long-horizon complex task, an agent must acquire observations, revise plans, invoke tools, verify results, and, after a local failure, decide whether to retry, recover, or stop. Interleaving reasoning and action allows agents to adjust their behavior based on environmental feedback \cite{yao2022react}, while interactive evaluation has begun to expand from short processes to cross-page operations and stateful business workflows \cite{liu2023agentbench,ma2024agentboard,zhou2023webarena,yao2024taubench}, as well as the simulated year-long business management studied by YC-Bench \cite{he2026ycbench}. In these tasks, stopping prematurely may leave unverified artifacts, whereas repeated planning and ineffective calls consume resources needed later; moreover, a state change caused by one action may invalidate subsequent steps in the original plan. Observation granularity, recovery methods, and budget allocation therefore form interdependent control relationships, and the final outcome depends on the entire execution loop rather than only on single-step generation quality.

Existing studies use external memory, linguistic reflection, and experience extraction to reuse past events or failures in subsequent decisions \cite{park2023generativeagents,shinn2023reflexion,zhao2024expel}; executable skill libraries further support the retrieval and composition of procedural behaviors \cite{wang2024voyager}. Meanwhile, declarative pipelines, code-based workflows, and modular agents bring structures external to the model into automated optimization \cite{khattab2024dspy,hu2024adas,zhang2025aflow,shang2025agentsquare}, and trajectory feedback has also been used to continually revise multi-step reasoning processes \cite{guo2025seagent}. These directions improve experiential content and execution structure, respectively, but they have not represented the overall control relationships among observation, planning, memory use, verification, recovery, stopping, and budget allocation as a unified object. Memory can provide content from the past, and skills can reproduce local operations, but neither inherently specifies when to invoke experience, how to check execution results, or when to terminate. Although historical traces can be stored or retrieved, their successes and failures remain difficult to convert into comparable and transferable modifications to the complete loop.

The recently emerging perspective of Loop Engineering treats triggers, goals, verification, stopping rules, and memory organization as design elements of loops external to the model \cite{macedo2026loopengineering}, providing a new level of analysis for the above problem. A loop object that can accumulate across runs also requires stable control semantics, host interfaces, and version relationships so that the same modification has comparable meaning across different runs. Loop optimization must also separate candidate generation from formal release: immediately rewriting the loop without candidate validation may introduce incidental noise or tail degradation into subsequent tasks, whereas always retaining a fixed loop prevents the system from learning from recurring failures. Research on evidence gating has highlighted the distinction between generated claims and verification states \cite{huang2026prooforstop}; Champion--Challenger mechanisms also constrain version replacement through candidate comparisons in predictive-model management and online AutoML \cite{nath2007champion,wu2021chacha}. However, these mechanisms have not been applied directly to complete agent loops, nor do they provide a shared policy asset and release semantics for offline historical trajectories and online operational feedback.

Against this background, this paper focuses on two related questions: first, how to separate the complete execution loop from the host implementation and a single context to form an external policy asset that can be versioned, ported, compared, and reused; and second, how to autonomously improve this asset using historical traces and current feedback while allowing only policies that have passed candidate validation, been released, and been activated at a task boundary to affect future runs. Online updates need to avoid disrupting the state of a task currently under execution, whereas offline search needs to form comparable candidate evaluations from multiple heterogeneous traces. The two modes thus differ in when evidence becomes available but require consistent asset and release semantics. The first question determines the object that serves as the unit of loop evolution, and the second determines how evidence drives policy changes and when those changes can be released and activated. The objective of this paper is not to allow an agent to rewrite itself arbitrarily within a single run, but to place policy search outside the execution path so that candidates, evaluation evidence, and formal versions remain clearly separated.

To address these questions, this paper presents OpenLoopEvolve (OLE), a verifiable self-evolution framework for Loop Policies in long-horizon complex tasks. OLE organizes observation, planning, memory, action, verification, recovery, stopping, and budget control into an externalized Loop Policy, and encapsulates the policy, its applicability conditions, and its version provenance in a Bundle within a policy lineage. The Loop Policy constrains a run through general control semantics without binding it to the specific model and tool implementation of the host, enabling loop experience to be read, compared, and reused independently of the original trace. A run's trace is then distilled into traceable evidence that points to policy modifications, after which a large language model autonomously proposes candidates.

OLE provides online and offline evolution modes according to the evidence source. The online mode generates a single Challenger based on recent feedback; following paired evaluation and a robust gate, the Challenger is released as a successor asset, activated at a subsequent task boundary, and then subject to degradation monitoring and rollback. The offline mode constructs a candidate population from archived traces and performs multi-generation search through batch evaluation, elite retention, and autonomous mutation and recombination. The two modes can be adopted separately according to run continuity, budget, and candidate-validation timing, while sharing the boundary of ``evidence--candidate--candidate validation--release--activation'': candidate generation belongs to the search process, only policies satisfying the candidate-validation requirements can be released, and only a released policy activated at a task boundary can enter the execution path. This forms a closed loop from historical interactions to subsequent policies, as illustrated in Fig.~\ref{fig:openloopevolve-framework}, allowing control experience to evolve continually while preventing provisional candidates from directly altering the execution path.

The main contributions of this paper are summarized as follows:
\begin{enumerate}
    \item We propose the Loop Policy and Bundle protocol, which represents complete loop control as an external policy asset with explicit interfaces, versions, and lineages, enabling historical control experience to be preserved, compared, exchanged, and reused across runs.
    \item We propose online and offline self-evolution methods based on trace evidence. A large language model autonomously proposes candidates, while paired evaluation, robust release, task-boundary activation, and rollback govern their entry into the execution path. To our knowledge, this work is the first to apply the Champion--Challenger release mechanism to both offline and online self-evolution of externalized Loop Policies.
    \item We evaluate OLE on YC-Bench. Results obtained with fixed models, benchmark configurations, and official seeds show that both online and offline modes outperform the fixed initial Loop Policy in long-term returns, task success rate, annual survival, and risk metrics.
\end{enumerate}

The remainder of this paper is organized as follows. Section II reviews research on long-horizon interactive agents, reuse of memory and trajectory experience, agent workflow optimization, and self-evolution. Section III defines long-horizon complex tasks, the loop interaction process, the trace, evolution evidence, Loop Policy, and Bundle. Section IV introduces the shared evolution framework of OpenLoopEvolve and its online and offline implementations. Section V presents the experimental results and analysis. Section VI concludes the paper. The code for this paper can be found at \url{https://github.com/yoyoshikc/OpenLoopEvolve}.

\section{Related Work}

\subsection{Long-Horizon Interactive Agents and Loop Execution}

Multi-round interaction with an environment is a fundamental capability of large language model agents in complex tasks. Unlike one-shot answer generation, an agent must receive environment states, select actions, and use new feedback to advance its objective over a sequence of decisions; task performance therefore depends on consistent reasoning and execution across steps. Existing comprehensive benchmarks have examined these capabilities in diverse interactive environments and have further recorded agents' intermediate progress before task completion~\cite{liu2023agentbench,ma2024agentboard}.

One line of research on this process constructs an execution loop in which reasoning and action are interleaved. Yao et al.~\cite{yao2022react} organize language reasoning, external actions, and environment observations within the same trajectory, enabling the model to adjust subsequent judgments based on execution outcomes. Another line of research embeds such loops in workflows involving real state changes: Zhou et al.~\cite{zhou2023webarena} evaluate cross-page operations and functional correctness in web environments, while Yao et al.~\cite{yao2024taubench} further evaluate task completion in interactions involving tools, users, and domain states. Correspondingly, long-horizon benchmarks have also begun to emphasize the cumulative consequences of control decisions: He et al.~\cite{he2026ycbench} test long-term planning and consistent execution through simulated business operations sustained over a year, rather than evaluating only a single-round response.

These studies establish the closed-loop interaction paradigm for long-horizon tasks and demonstrate the roles of planning, tool use, and state feedback in task completion. Their primary objective, however, remains to improve or evaluate agent behavior within a given execution structure; how observations are organized, when planning or recovery occurs, how outcomes are verified, and when execution stops are typically determined by the particular method or host framework. This leaves room to investigate how these control rules can be separated from the implementation details of an individual run and treated as a unified object for comparison and optimization. Enabling loop control rules to improve continually across runs also requires addressing how historical interactions can be accumulated and brought back into subsequent decisions.

\subsection{Memory, Reflection, and Reuse of Trajectory Experience}

Memory and reflection provide the main means for agents to accumulate experience across rounds. Their shared premise is to preserve historical interactions beyond the limited context window in an external medium and retrieve, summarize, or reorganize them during subsequent decisions, thereby maintaining consistency in long-term behavior. Park et al.~\cite{park2023generativeagents} connect the recording of experiences with reflection and planning, enabling agents to derive higher-level insights from past events and use them to guide subsequent behavior; this demonstrates that historical trajectories can contribute to continual decision-making beyond serving as raw conversation records.

Existing methods differ in how they preserve and use experience. Shinn et al.~\cite{shinn2023reflexion} convert task feedback into linguistic reflections and store them in episodic memory, allowing subsequent trials to draw on failures without updating model parameters; Zhao et al.~\cite{zhao2024expel} extract natural-language experiences from a set of training tasks and jointly retrieve abstract insights and similar successful trajectories at inference time. In contrast to textual experience, Wang et al.~\cite{wang2024voyager} store successful programs generated from environment feedback in a retrievable skill library, allowing procedural behaviors to be composed and reused in new Minecraft worlds. These methods show, respectively, that historical trajectories can be distilled into reflective text, experiential knowledge, or executable skills.

These mechanisms substantially expand agents' ability to reuse past experience, but their update targets are primarily memory content, linguistic experience, or task skills, and they generally do not directly represent the overall control relationships among observation, planning, verification, recovery, stopping, and budget allocation. This paper does not seek to add another form of memory content; instead, it distills evidence from traces that supports policy comparison and uses this evidence to modify how the complete loop is controlled. This distinction leads further to the question of agent workflows and their self-evolution.

\subsection{Agent Workflow Optimization and Self-Evolution}

As agent systems have expanded from single prompts into composite programs containing model calls, tools, and control flow, the object of study has gradually shifted from prompt optimization to the optimization of workflows and agent structures. Khattab et al.~\cite{khattab2024dspy} represent language model pipelines as composable declarative modules and compile their parameters against specified metrics; Hu et al.~\cite{hu2024adas} instead represent agent systems as code, enabling a meta-agent to generate and search for new agent programs, with a code space that can in principle cover prompts, patterns of tool use, and workflow compositions. These studies show that execution structures external to the model can themselves become objects of automatic optimization.

In terms of the evidence used for optimization, existing work can be summarized along two paths: offline structural search and continual improvement driven by runtime feedback. Offline search uses existing tasks, execution evaluations, and accumulated search experience to produce candidate structures: Zhang et al.~\cite{zhang2025aflow} model code-represented Agent Workflows as a search space and iteratively modify workflows using execution feedback and tree-structured experience; Shang et al.~\cite{shang2025agentsquare} abstract planning, reasoning, tool use, and memory into modules with unified interfaces and search for agent designs through evolution and recombination. By comparison, trajectory-driven self-evolution research places greater emphasis on using newly generated interaction feedback to adjust subsequent solution processes; Guo et al.~\cite{guo2025seagent} extract feedback from multi-step reasoning trajectories and iteratively optimize the reasoning process. The former focuses on searching for structures in controlled evaluation environments, whereas the latter focuses on allowing runtime experience to continually influence subsequent behavior; its focus nevertheless remains iterative optimization of the reasoning process and does not yet address policy release governance for deployment. Together, these two paths move agents from fixed designs toward experience-driven continual optimization.

The recent emergence of Loop Engineering further concentrates attention on loop structures external to the model. Macedo~\cite{macedo2026loopengineering} organizes trigger conditions, objectives, verification, stopping rules, and memory into reusable loop specifications, while Huang et al.~\cite{huang2026prooforstop} emphasize that only evidence satisfying a verification gate can drive state transitions in the agent lifecycle. These studies extend workflow optimization to the systematic design of complete execution loops and begin to address verification and state governance within them. In related work, the Champion--Challenger paradigm has been used for the continual comparison and replacement of production predictive models~\cite{nath2007champion} and has been further developed into a hyperparameter-selection algorithm that concurrently maintains one Champion and multiple online Challengers~\cite{wu2021chacha}. Existing methods, however, primarily compare predictive models or learning configurations. Loop Engineering itself has not yet established unified object boundaries and interface standards; most existing optimization results remain specific programs, task structures, or configurations internal to a framework, while offline search and online runtime feedback still lack a shared policy asset and release semantics. To address this gap, this paper represents loop control logic as a versioned and portable Loop Policy, uses a unified protocol to accommodate both offline and online candidates, and governs their evolution through Champion--Challenger comparison, evidence-constrained release, and rollback. To the best of our knowledge, this paper is the first to apply a Champion--Challenger release mechanism to the offline and online self-evolution of an externalized Loop Policy, enabling loop-control experience from long-horizon tasks to be verified, released, and reused as an independent asset.

\section{Problem Formulation}
\label{sec:problem-definition}

\subsection{Long-Horizon Tasks and the Loop-Control Problem}

Sustained execution of multi-round tasks in interactive environments is an important test of an agent's reasoning, decision-making, and execution capabilities, and has become a major focus of general-agent research and evaluation~\cite{liu2023agentbench,ma2024agentboard}. Such tasks require an agent to engage in closed-loop interaction with an external environment over successive decision rounds toward a given objective: the agent selects an action based on its current observation, the action changes the environment state and produces new feedback, and subsequent decisions therefore depend on prior interactions. Existing research characterizes the long-horizon nature of these tasks primarily at two levels: solution mechanisms and evaluation protocols. At the solution-mechanism level, Yao et al.~\cite{yao2022react} organize task execution as closed-loop interaction, enabling an agent to update its plan in response to feedback and handle exceptions. At the evaluation-protocol level, WebArena and Tau-bench~\cite{zhou2023webarena,yao2024taubench} construct tasks as realistic interactive workflows with multistep dependencies and determine task completion from functional correctness or the post-execution environment state. The former reflects closed-loop decision-making and state dependence during execution, whereas the latter ensures that task outcomes can be verified through the external environment.

Following the closed-loop interaction perspective, we view the solution of a long-horizon complex task as an execution loop in which an agent repeatedly cycles through reasoning, action, and observation of the environment, and we focus on the mechanisms that control observation, planning, action, verification, recovery, and stopping within this loop. We define a long-horizon complex task as one whose solution process cannot be reliably reduced to a single model generation without environment interaction, but instead requires an agent to execute multiple consecutive decision steps with state dependencies under finite resource constraints, adjusting subsequent behavior according to tool returns and environment feedback until it produces a verifiable task outcome. Formally, we describe a long-horizon complex task using the following executable contract:
\begin{equation}
g=\langle d,\mathcal{U}_g,\mathcal{M}_g,\mathcal{S}_g,\mathcal{V}_g,\boldsymbol{\beta}_g\rangle,
\qquad g\in\mathcal{G}.
\label{eq:task-contract}
\end{equation}
Here, $\mathcal{G}$ denotes the space of long-horizon complex tasks. $d$ describes the task objective, expected artifact, and initial conditions, providing an anchor for planning and replanning within the loop. $\mathcal{U}_g$ specifies the set of executors permitted by the task, thereby restricting the tool calls and environment actions available to the loop. $\mathcal{M}_g$ provides the evaluation metrics that determine how execution progress, outcome quality, and resource cost are measured. $\mathcal{S}_g$ restricts the evidence sources that may support a judgment, such as environment observations, run logs, and task artifacts, preventing task completion from being established solely by the model's own statements. $\mathcal{V}_g$ specifies when and how verification is performed and feeds its results back into replanning, recovery, and stopping. $\boldsymbol{\beta}_g$ is the resource-budget vector comprising upper bounds on quantities such as interaction rounds, tool calls, time, and tokens, and thus bounds retries and exploration. These six fields specify the loop's objective, capability boundary, measurement criteria, grounds for judgment, verification mechanism, and resource boundary, respectively.

Based on this formal definition, we further characterize an agent's loop interaction process within a single run to make explicit the dynamic relationships among observations, internal states, actions, and environment feedback. Before giving its structured components, we use $\Pi$ to denote the Loop Policy space abstractly and let $\pi\in\Pi$ denote the Loop Policy that constrains the current execution loop; its eight-component structure is defined in Section~\ref{subsec:loop-policy-definition}. The loop-control process can be abstracted as
\begin{equation}
x_{t+1}=f_{\pi}(x_t,o_t,a_t,e_t),\qquad
a_{t+1}\sim h_{\pi}(x_{t+1}),
\label{eq:state-transition}
\end{equation}
where $t$ denotes the current interaction step; $x_t\in\mathcal{X}$ denotes the internal control state at step $t$; $o_t\in\mathcal{O}$ denotes the observation that the agent receives from a tool or the environment at step $t$; $a_t\in\mathcal{A}$ denotes the action executed at step $t$; $\mathcal{E}_{\mathrm{env}}$ denotes the environment-event space, and $e_t\in\mathcal{E}_{\mathrm{env}}$ denotes the environment event caused by action $a_t$; $\pi$ denotes the Loop Policy that constrains loop behavior; $f_{\pi}$ denotes the update function that, under Loop Policy $\pi$, computes the successor state from $x_t$, $o_t$, $a_t$, and $e_t$; $x_{t+1}$ denotes the next control state produced by $f_{\pi}$; $h_{\pi}$ denotes the decision function that selects an action from $x_{t+1}$; and $a_{t+1}$ denotes the next action produced by $h_{\pi}$. An action may be a tool call, information acquisition, artifact modification, outcome verification, or stopping decision, and its available range is jointly constrained by $\mathcal{U}_g$ in the task contract and the host's capabilities. The state $x_t$ is an abstract representation of the control information required to run the loop and need not be a fully observable Markov state. Factors including observation, planning, memory use, reflection, risk handling, evaluation, and budget allocation affect subsequent state updates and action selection through $f_{\pi}$ and $h_{\pi}$.

Having characterized the loop interaction process, we must also specify the basis for evaluating a Loop Policy and its optimization objective. Let $q(\pi;g,z)\in\mathbb{R}$ denote the value score attained by Loop Policy $\pi$ on task $g$ under controlled run condition $z\in\mathcal{Z}$, where $\mathcal{Z}$ denotes the run-condition space and $q$ is determined by the evaluation metrics $\mathcal{M}_g$ and verification protocol $\mathcal{V}_g$ in the task contract. Under these conventions, Loop Policy optimization seeks, for the same task and controlled run condition, a Loop Policy with a higher value score that satisfies the verification requirements and resource boundary. This objective can be formalized as
\begin{equation}
\begin{aligned}
\pi^{\star}\in\arg\max_{\pi\in\Pi}\quad &q(\pi;g,z)\\
\text{s.t.}\quad
&\mathbf{c}_{\mathcal{V}_g}(\pi;g,z)
\preceq\boldsymbol{\kappa}_g,\\
&\mathbf{b}(\pi;g,z)\preceq\boldsymbol{\beta}_g,
\end{aligned}
\label{eq:loop-policy-optimization}
\end{equation}
where $\Pi$ is the Loop Policy space formally defined in Section~\ref{subsec:loop-policy-definition}, $\pi^\star$ denotes an optimal Loop Policy for this constrained optimization problem, $\mathbf{c}_{\mathcal{V}_g}(\pi;g,z)$ denotes the verification-risk vector determined by verification protocol $\mathcal{V}_g$, $\boldsymbol{\kappa}_g$ denotes the corresponding admissible upper bounds, $\mathbf{b}(\pi;g,z)$ denotes the resource-consumption vector, and $\preceq$ denotes elementwise vector comparison. This optimization problem jointly characterizes outcome quality, verification reliability, and resource boundaries. The observations, actions, and environment feedback produced during Loop Policy execution further form interaction traces, providing empirical grounds for analyzing existing Loop Policy behavior and supporting subsequent evolution. The next subsection accordingly defines a trace and its evidence form.

\subsection{Traces and the Evidence Contract}

The continual evolution of a Loop Policy depends on the effective use of historical runs, requiring the execution process to be formalized as a trace that links the task, Loop Policy, and outcome. Let the trace of the $i$th run be
\begin{equation}
\tau_i=(g_i,\pi_{v(i)},\mathcal{H}_i,y_i),
\qquad
\mathcal{H}_i=\{(o_t,a_t,e_t)\}_{t=1}^{T_i},
\label{eq:trace}
\end{equation}
where $g_i$ denotes the task processed in the $i$th run, $v(i)$ denotes the stable version index of the registered Bundle used in that run, and $\pi_{v(i)}$ denotes the Loop Policy in that Bundle. The stable index $v(i)$ can be used to retrieve the complete asset $\mathcal B_{v(i)}$, including its applicability conditions and version provenance. $\mathcal{H}_i$ denotes the interaction history composed of observations, actions, and environment events, $T_i$ denotes the trace length, and $y_i$ denotes the run outcome. This representation explicitly preserves the mapping between policy-asset identity and the corresponding interaction outcome, ensuring that historical performance can be accurately attributed to a specific registered Bundle version.

After obtaining attributable traces, we must further distill from them evolution evidence that can guide adjustments to the Loop Policy. We define such evidence as a testable basis for modifying a specific Loop Policy component, inferred from a set of traces that are attributable, whose outcomes are verifiable, and whose evidence sources conform to the task contract. Each evidence item must specify the recommended direction of adjustment, the associated performance change, the policy component concerned, and the original traces supporting the judgment. Formally, the evidence-extraction operator $\Phi$ maps a trace set $\mathcal{D}$ to
\begin{equation}
\mathcal{E}=\Phi(\mathcal{D})
=\{\varepsilon_j=(\rho_j,\delta_j,u_j,\mathcal{I}_j)\}_{j=1}^{J},
\label{eq:evidence-extraction}
\end{equation}
where $\varepsilon_j$ denotes the $j$th item of evolution evidence and $J$ denotes the number of evidence items; $\rho_j\in\{+,-\}$ denotes the direction of the evidence: positive evidence indicates that the relevant behavior pattern is consistent with improved performance and may support Loop Policy reuse, whereas negative evidence indicates that the relevant behavior pattern is consistent with failure or performance degradation and may support Loop Policy revision; $\delta_j$ denotes a metric change relative to a control run or a statistical reference for comparable traces; $u_j$ denotes the Loop Policy component targeted by the evidence, taking a value corresponding to observation, planning, memory, action, verification, recovery, stopping, or budget control as defined in the next subsection; and $\mathcal{I}_j\subseteq\{i\mid\tau_i\in\mathcal D\}$ denotes the set of global trace indices supporting the judgment.

In summary, the trace and evidence contract transforms historical interactions into attributable and verifiable evolution evidence that can be used to adjust the Loop Policy, laying the foundation for defining the modifiable and reusable Loop Policy object in the next subsection.

\subsection{Loop Policy}
\label{subsec:loop-policy-definition}

Run evidence must act on an explicit loop-control object before it can be converted into reusable Loop Policy experience. Loop Engineering extends the scope of engineering to the external agent loop and systematically designs its objectives, feedback, verification, stopping, and cross-run experience~\cite{macedo2026loopengineering}. Building on the preceding abstract type declaration, we define the set of external rules that controls how an agent advances through a complete execution loop as a Loop Policy, whose eight-component structure is denoted by
\begin{equation}
\begin{aligned}
\pi=\langle
&\pi_{\mathrm{obs}},\pi_{\mathrm{plan}},\pi_{\mathrm{mem}},\pi_{\mathrm{act}},\\
&\pi_{\mathrm{ver}},\pi_{\mathrm{rec}},\pi_{\mathrm{stop}},\pi_{\mathrm{bud}}
\rangle\in\Pi,
\end{aligned}
\label{eq:loop-policy}
\end{equation}
where $\Pi$ denotes the Loop Policy space; $\pi_{\mathrm{obs}}$ governs the acquisition and filtering of observations, $\pi_{\mathrm{plan}}$ governs plan generation and updating, $\pi_{\mathrm{mem}}$ governs memory writing, retrieval, and use, $\pi_{\mathrm{act}}$ governs action and tool selection, $\pi_{\mathrm{ver}}$ governs outcome verification, $\pi_{\mathrm{rec}}$ governs recovery from failure, $\pi_{\mathrm{stop}}$ governs the loop's stopping conditions, and $\pi_{\mathrm{bud}}$ governs resource-budget allocation and adjustment. Together, these eight components determine the state-update relation $f_\pi$ and action-decision relation $h_\pi$ in Eq.~\eqref{eq:state-transition}, allowing a Loop Policy to cover the complete decision process from incorporating environment feedback into the control state to producing the next action.

\begin{figure}[!t]
\centering
\includegraphics[width=0.94\columnwidth]{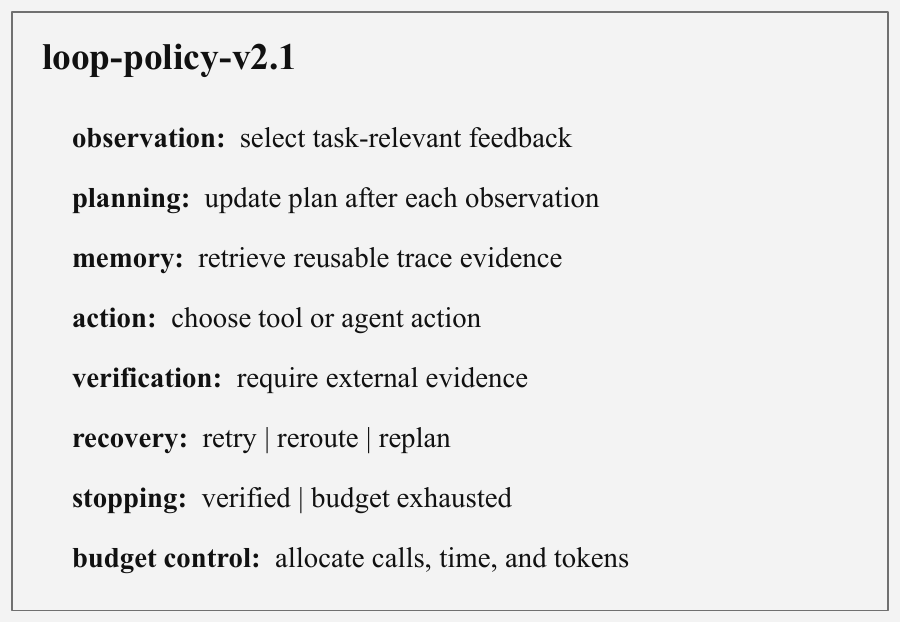}
\caption{A YAML-style example of a Loop Policy. The boldface fields correspond to the eight categories of loop-control rules: observation, planning, memory, action, verification, recovery, stopping, and budget control.}
\label{fig:loop-policy-example}
\end{figure}

Fig.~\ref{fig:loop-policy-example} presents a simplified Loop Policy example in a YAML-style format. This representation organizes the eight categories of loop rules into a readable and modifiable external structure, allowing Loop Policy content to be stored, compared, and reused independently of any single-run context.

To support the reuse of loop-control experience across tasks, we design a Loop Policy as an external asset independent of a specific host agent and use a unified representation to carry its control semantics. A Loop Policy can be integrated as a pluggable object into different agent frameworks through an adaptation interface, without binding its content to a particular host implementation, thereby supporting Loop Policy replacement, comparison, and evolution across agent frameworks. This externalized design provides the basis for the versioned asset representation in the next subsection.

\subsection{Asset Representation of a Loop Policy}

As a control object that can be independently compared and reused, a Loop Policy also requires a stable asset representation. Reusing a Loop Policy across runs requires the Loop Policy itself, its applicability conditions, and its version provenance to be jointly identifiable. We represent the asset corresponding to the $v$th registered Bundle version as a Bundle:
\begin{equation}
\mathcal{B}_v=\langle \pi_v,\chi_v,\ell_v\rangle,
\label{eq:bundle}
\end{equation}
where $v$ is the stable version index of a registered Bundle, $\pi_v$ denotes the Loop Policy of this version, $\chi_v\subseteq\mathcal{Z}$ denotes the set of applicability conditions for this Loop Policy, and $\ell_v$ denotes the version provenance, including the stable index $p_v$ of the direct parent version and the evolution evidence on which the version is based; the initial Bundle has $p_v=\bot$ and no supporting evolution evidence. Together, these three components describe which Loop Policy is used, under what conditions it is executed, and its version provenance, allowing the Loop Policy to be stored and reused as a complete object. Registration assigns a stable index when a valid candidate Bundle is formed so that it can enter evaluation history and lineage; registration itself does not mean that the candidate has passed the gate, been released, or been activated. Hereafter, $k$ is reserved for evolution rounds or offline generation indices and no longer serves as a stable version identity.

Bundles are compared under shared applicability conditions $z\in\chi_{\mathrm{par}}\cap\chi_{\mathrm{cand}}$, where the subscripts $\mathrm{par}$ and $\mathrm{cand}$ denote the parent and candidate versions, respectively, and $\pi_{\mathrm{par}}$ and $\pi_{\mathrm{cand}}$ are the corresponding Loop Policies. For each noninitial version with $p_v\neq\bot$, the version provenance $\ell_v$ links it to its direct parent version and to the run evidence supporting the change. The directed relationships from parent versions to successor versions form a policy lineage, allowing historical experience to accumulate along version relationships and supporting subsequent candidate comparison, version selection, and outcome tracing.

This section has defined, in sequence, the loop interaction process, traces and evolution evidence, the Loop Policy, and its asset representation, beginning with the long-horizon complex task. These definitions specify the objects of self-evolution, the sources of experience, and the basis for comparison, providing a unified notation for the next section's presentation of online and offline evolution, the candidate lifecycle, and the robust release gate.

\section{OpenLoopEvolve Method}

Section III formally defined long-horizon complex tasks, traces, evolution evidence, and Loop Policies. Building on these definitions, OpenLoopEvolve converts traces accumulated across runs into a basis for modifying Loop Policies and optimizes an external Loop Policy rather than the parameters of the base model. Starting from the current Champion, the method constructs evidence, generates candidates, performs paired evaluation, and applies constrained release to produce a successor version. Depending on the evidence source and update timing, this common framework can be instantiated as either online or offline evolution.

\subsection{Overall OpenLoopEvolve Framework}

The two evolution modes of OpenLoopEvolve share the same processing chain from run experience to policy assets. Let $\mathcal{D}_k$ denote the set of traces used in evolution round $k$. The corresponding evidence-construction process is
\begin{equation}
\mathcal{E}_k=\Phi(\mathcal{D}_k),\qquad
\mathcal{D}_k\in
\{\mathcal{D}^{\mathrm{off}},\mathcal{D}^{\mathrm{on}}_k\},
\label{eq:evolution-modes}
\end{equation}
where $\Phi$ follows Eq.~\eqref{eq:evidence-extraction}; $\mathcal D^{\mathrm{off}}$ and $\mathcal D_k^{\mathrm{on}}$ are attributable traces available from archives and online operation, respectively. The online mode retains the round index and uses $\mathcal E_k=\Phi(\mathcal D_k^{\mathrm{on}})$; because the offline archive remains fixed across generations, $\Phi(\mathcal D^{\mathrm{off}})$ is abbreviated as $\mathcal E$ below. A trace can enter $\mathcal D_k$ only if its outcome has been verified by $\mathcal V_{g_i}$ and its source belongs to $\mathcal S_{g_i}$. Candidate generation reads $\varepsilon_j=(\rho_j,\delta_j,u_j,\mathcal I_j)$, where $u_j$ identifies a component in Eq.~\eqref{eq:loop-policy}; $G^{\mathrm{on}}_{\mathrm{LLM}}$ returns one registered Bundle, whereas $G^{\mathrm{off}}_{\mathrm{LLM}}$ and $V^{\mathrm{off}}_{\mathrm{LLM}}$ return sets of registered Bundles; a proposal that fails to form a valid Bundle yields the empty value or is excluded from the set. A candidate may be released as a successor asset only after passing paired evaluation and the gate. The host activates a released policy only at a task boundary, and only the activated policy controls environment interaction according to Eq.~\eqref{eq:state-transition} and produces new traces as defined in Eq.~\eqref{eq:trace}, thereby forming the closed loop shown in Fig.~\ref{fig:openloopevolve-framework}.

For readability, Table~\ref{tab:main-notation} summarizes the key symbols used throughout the problem formulation and method in the order of task and interaction, trace and evidence, policy assets, and evolution and release. Indices, temporary variables, and hyperparameters used only within a single equation or algorithm are defined locally.

\begin{table*}[!t]
\caption{Main notation and meanings}
\label{tab:main-notation}
\centering
\small
\renewcommand{\arraystretch}{1.05}
\begin{tabular}{p{0.13\textwidth} p{0.28\textwidth} p{0.52\textwidth}}
\toprule
Category & Symbol & Meaning \\
\midrule
\multirow{3}{*}{\shortstack{Task and\\ interaction}}
& $g=\langle d,\mathcal U_g,\mathcal M_g,\mathcal S_g,\mathcal V_g,\boldsymbol\beta_g\rangle$ & Executable contract for a long-horizon task, comprising its objective, permitted executors, evaluation metrics, evidence sources, verification protocol, and resource budget \\
& $x_t,o_t,a_t,e_t$; $f_\pi,h_\pi$ & Control state, observation, action, and environment event at step $t$; state-update and action-decision functions constrained by the policy \\
& $z\in\mathcal Z$; $q(\pi;g,z)$; $\mathbf c_{\mathcal V_g},\boldsymbol\kappa_g$; $\mathbf b,\boldsymbol\beta_g$ & Controlled run condition; value score; verification risk and its upper bound; resource consumption and its upper bound \\
\midrule
\multirow{3}{*}{\shortstack{Traces and\\ evidence}}
& $\tau_i=(g_i,\pi_{v(i)},\mathcal H_i,y_i)$ & Attributable trace of run $i$, including the task, stable Bundle version, interaction history, and run outcome \\
& $\mathcal D,\mathcal E=\Phi(\mathcal D)$ & Set of attributable traces with verified outcomes and compliant evidence sources; the evolution-evidence set; and the evidence-extraction operator \\
& $\varepsilon_j=(\rho_j,\delta_j,u_j,\mathcal I_j)$ & Direction, metric change, target policy component, and supporting trace indices of evolution-evidence item $j$ \\
\midrule
\multirow{3}{*}{Policy assets}
& $\pi\in\Pi$ & Loop Policy and its policy space; the policy comprises rules for observation, planning, memory, action, verification, recovery, stopping, and budget control \\
& $\mathcal B_v=\langle\pi_v,\chi_v,\ell_v\rangle$ & Bundle corresponding to stable version index $v$, comprising a Loop Policy, a set of applicability conditions, and version provenance \\
& $\chi_v,\ell_v$ & Bundle applicability conditions and version provenance containing the parent version and supporting evidence \\
\midrule
\multirow{7}{*}{\shortstack{Evolution and\\ release}}
& $\mathcal D^{\mathrm{off}},\mathcal D_k^{\mathrm{on}}$; $\mathcal E,\mathcal E_k$ & Archived offline traces, traces visible in online round $k$, and their corresponding evolution evidence \\
& $G^{\mathrm{on}}_{\mathrm{LLM}},G^{\mathrm{off}}_{\mathrm{LLM}},V^{\mathrm{off}}_{\mathrm{LLM}}$ & Online single-candidate generation, offline initial-candidate-set generation, and offline candidate mutation and recombination operators \\
& $\mathcal R=\mathcal R^{\mathrm{on}}\uplus\mathcal R^{\mathrm{off}}$ & Mode-typed evaluation histories; the online component records candidates, evaluations, and release decisions, whereas the offline component records candidate sets and evaluation mappings \\
& $\mathcal B_c,\mathcal B_{k,j}\equiv\mathcal B_{v_{k,j}}$ & Champion and Challenger Bundle $j$ in round $k$; $v_{k,j}$ is its stable version index (online: $j=1$, $\mathcal B'_k\equiv\mathcal B_{k,1}$) \\
& $\Delta_{i,j},r_{i,j},\mathcal P_{k,j}$ & Absolute and relative value-score changes for pair $i$, and the complete paired evaluation of a candidate \\
& $\Gamma_\alpha,\Theta_\alpha,\operatorname{Gate}$ & Release constraint of category $\alpha$, its configuration, and the aggregate release-gate operator \\
& $\mathcal C_k,\mathcal F_k,\overline{\mathcal P}_k$ & Candidate set, retained set comprising the fixed Champion and selected nondominated candidates, and cumulative paired-evaluation mapping at offline generation $k$ \\
\bottomrule
\end{tabular}
\end{table*}

Champion--Challenger paired evaluation provides a common comparison basis for both modes~\cite{nath2007champion,wu2021chacha}. Let the current Champion Bundle be $\mathcal B_c=\langle\pi_c,\chi_c,\ell_c\rangle$, and let candidate Bundle $j$ generated in round $k$ be $\mathcal B_{k,j}=\langle\pi_{k,j},\chi_{k,j},\ell_{k,j}\rangle\equiv\mathcal B_{v_{k,j}}$, where $(k,j)$ is the candidate's search coordinate and $v_{k,j}$ is the stable version index assigned when the candidate is formed. The two are executed separately only on the same task $g_i$ and under the same controlled run condition $z_{i,j}\in\chi_c\cap\chi_{k,j}$. For pair $i$ of candidate $j$, the value-score difference and relative change are defined as
\begin{equation}
\begin{aligned}
\Delta_{i,j}
&=q(\pi_{k,j};g_i,z_{i,j})-q(\pi_c;g_i,z_{i,j}),\\
r_{i,j}
&=\frac{\Delta_{i,j}}
{\max\{|q(\pi_c;g_i,z_{i,j})|,\epsilon\}},
\end{aligned}
\label{eq:paired-gain}
\end{equation}
where $\Delta_{i,j}$ is the candidate's absolute value-score change relative to the Champion, $r_{i,j}$ is the relative change, and $\epsilon>0$ ensures numerical stability when the Champion value score is close to zero. In each paired run, the Loop Policy in the corresponding Bundle controls the execution loop in Eq.~\eqref{eq:state-transition}. The value score $q$ remains determined by $\mathcal M_{g_i}$ and $\mathcal V_{g_i}$ in the task contract, while the action scope and resource upper bounds comply with $\mathcal U_{g_i}$ and $\boldsymbol\beta_{g_i}$, respectively. Holding the task, model, tools, and resource conditions constant controls for the primary external sources of variation, allowing paired changes to reflect the effect of Loop Policy modifications more directly. The valid relative changes for candidate $j$ form the sample $r_j=\{r_{i,j}\}$ and, together with the verification risk $\mathbf c_{\mathcal V_{g_i}}$ in Eq.~\eqref{eq:loop-policy-optimization}, resource consumption $\mathbf b$, failure states, and evidence validity, constitute the paired evaluation $\mathcal P_{k,j}$. For runs with termination failures or invalid evidence, the evaluation protocol uses a bounded failure-aware change and retains the corresponding validity and reliability states. Thus, $\operatorname{PairEval}(\mathcal B_c,\mathcal B_{k,j})$ in the algorithms below always denotes paired evaluation of complete Bundles under shared applicability conditions, rather than a comparison of two bare policies without regard to applicability conditions.

Paired evaluation does not directly replace the current policy according to a single mean score; it provides evidence for constrained release. Let $\Gamma_b$, $\Gamma_e$, $\Gamma_t$, and $\Gamma_c$ indicate whether the four constraint categories of benefit, evidence quality, tail risk, and resource cost are satisfied, respectively. The release gate can be expressed as
\begin{equation}
\begin{aligned}
&\operatorname{Gate}(\mathcal P_{k,j},\Theta)=\mathrm{accept}\\
&\quad\Longleftrightarrow\quad
\prod_{\alpha\in\mathcal J_{\mathrm{gate}}}
\Gamma_\alpha(\mathcal P_{k,j},\Theta_\alpha)=1,
\end{aligned}
\label{eq:robust-selection}
\end{equation}
where $\mathcal J_{\mathrm{gate}}=\{b,e,t,c\}$ corresponds to benefit, evidence quality, tail risk, and resource cost; $\Gamma_\alpha(\mathcal P_{k,j},\Theta_\alpha)\in\{0,1\}$ indicates whether the constraint of category $\alpha$ is satisfied; and $\Theta=\{\Theta_\alpha\}$ is the gate configuration. $\Gamma_e$ checks the evidence permitted by $\mathcal S_{g_i}$, $\Gamma_t$ enforces $\mathbf c_{\mathcal V_{g_i}}\preceq\boldsymbol\kappa_{g_i}$, and $\Gamma_c$ enforces $\mathbf b\preceq\boldsymbol\beta_{g_i}$. The number of pairs, win rate, confidence interval, task failures, and tail losses may be used as internal evaluation quantities. A Challenger is eligible for release only if all enabled constraints are satisfied, and the configured thresholds cannot relax the verification and resource boundaries of the task contract.

Within this common framework, the online and offline modes are suited to different forms of available run experience. When the task distribution continues to change or new feedback must be incorporated promptly, the online mode can perform staged updates. When archived traces cover the main run states of the target task, the offline mode can conduct a concentrated search over multiple policy directions. The application selects the mode according to the scope of available traces, feedback timeliness, and update requirements; the system does not switch modes automatically. The modes differ in candidate-search procedures and update timing, while their policy comparison and release eligibility are governed uniformly by Eqs.~\eqref{eq:paired-gain} and~\eqref{eq:robust-selection}.

The evaluation histories of the two modes are explicitly typed. Let $\mathcal R=\mathcal R^{\mathrm{on}}\uplus\mathcal R^{\mathrm{off}}$ be a disjoint union. Each record in the online component $\mathcal R^{\mathrm{on}}$ is $(\mathcal B',\mathcal P,\omega)$, which stores a candidate Bundle, its paired evaluation, and the release decision, respectively. The record for offline generation $k$ in $\mathcal R^{\mathrm{off}}$ is $(\mathcal C_k,\mathcal P_k)$, which stores the candidate set and its evaluation mapping. Each generator reads only the history component of its own mode, preventing the two record structures from being interpreted as the same untyped tuple.

\subsection{Online Loop Policy Evolution}

Online evolution uses a continuous stream of recent feedback to update the Loop Policy in stages, allowing subsequent tasks to incorporate the latest run experience. At online update time $k$, the host submits completed traces attributable to explicit policy versions. Once the application-defined update condition is satisfied, these traces are converted into evolution evidence $\mathcal E_k$. The large language model then generates a Challenger Loop Policy and encapsulates it in a candidate Bundle:
\begin{equation}
\mathcal B'_k=G^{\mathrm{on}}_{\mathrm{LLM}}(\mathcal B_k,\mathcal E_k,\mathcal R^{\mathrm{on}}_{<k})
=\langle\pi'_k,\chi'_k,\ell'_k\rangle,
\label{eq:challenger-bundle}
\end{equation}
where $\mathcal B_k=\langle\pi_k,\chi_k,\ell_k\rangle$ is the current Champion held at online update time $k$, and $\mathcal R^{\mathrm{on}}_{<k}$ is the prior online evaluation history; here, $k$ denotes an update time, not a stable version index. The online mode has one candidate per round, for which we set $j=1$. $G^{\mathrm{on}}_{\mathrm{LLM}}$ rewrites $\pi'_k$, sets $\chi'_k\subseteq\chi_k$, and records the parent version and evidence from the current round in $\ell'_k$. When a valid candidate is formed, stable version index $v_{k,1}$ is assigned atomically and the candidate is written as $\mathcal B'_k\equiv\mathcal B_{k,1}\equiv\mathcal B_{v_{k,1}}$, thereby directly producing a registered candidate Bundle as defined in Eq.~\eqref{eq:bundle}. The candidate undergoes paired evaluation only within $\chi_k\cap\chi'_k$. If it passes the gate, it is recorded as a released successor asset and activated at a subsequent task boundary as $\mathcal B_{k+1}=\mathcal B'_k$; otherwise, $\mathcal B_{k+1}=\mathcal B_k$. To emphasize the processing sequence of a single update, Algorithm~\ref{alg:online-evolution} omits the update subscript $k$ within the loop body; $\mathcal B$, $\mathcal D^{\mathrm{on}}$, $\mathcal E$, and $\mathcal B'$ correspond respectively to $\mathcal B_k$, $\mathcal D_k^{\mathrm{on}}$, $\mathcal E_k$, and $\mathcal B'_k$ for the current batch.

Online updates must preserve policy consistency within each task and support rollback after release. A task is controlled by the same activated Champion from start to finish. A Challenger that passes the gate may be released but is activated only at a subsequent task boundary, so that run outcomes remain attributable to explicit policy versions. After the new version is activated, $\Call{CanaryMonitor}{}$ monitors its performance using attributable subsequent feedback. When the degradation condition is met, it rolls back to the parent version and quarantines the traces that triggered the rollback from the current update batch; other usable traces in the same batch may still participate in candidate generation. The update condition $\operatorname{Trig}$ may depend on the amount of feedback, an update period, or a task stage. Its only function is to determine when evolution begins; it does not replace the release gate in the overall framework. The online mode therefore forms a closed loop of feedback triggering, candidate validation, release, subsequent-task activation, and degradation rollback, as detailed in Algorithm~\ref{alg:online-evolution}.

\begin{olealgorithm}{Online Loop Policy Optimization}
\label{alg:online-evolution}
\begin{algorithmic}[1]
\Require Champion Bundle $\mathcal B=\langle\pi_c,\chi,\ell\rangle$, stream of attributable traces $\{\tau_i\}$ (with $y_i$ verified by $\mathcal V_{g_i}$ and its evidence source belonging to $\mathcal S_{g_i}$), update condition $\operatorname{Trig}$, online evaluation history $\mathcal R^{\mathrm{on}}$, gate configuration $\Theta$
\Ensure Champion $\mathcal B$ updated with the trace stream and online evaluation history $\mathcal R^{\mathrm{on}}$
\AlgExplain{Continuously receive batches of traces submitted by the host when checkpoint or feedback conditions are met.}
\While{new trace batches continue to arrive}
  \State $\mathcal D^{\mathrm{on}}\gets\Call{ReceiveTraces}{}$
  \AlgExplain{Monitor the performance of the new version; roll back a degraded version, and let $\widetilde{\mathcal D}^{\mathrm{on}}$ denote the usable traces after quarantining the corresponding feedback.}
  \State $(\mathcal B=\langle\pi_c,\chi,\ell\rangle,\widetilde{\mathcal D}^{\mathrm{on}})\gets\Call{CanaryMonitor}{\mathcal B,\mathcal D^{\mathrm{on}}}$
  \AlgExplain{Start evolution only when usable feedback exists and the external update condition is satisfied.}
  \State \textbf{if} $\widetilde{\mathcal D}^{\mathrm{on}}=\varnothing$ \textbf{or} $\neg\operatorname{Trig}(\widetilde{\mathcal D}^{\mathrm{on}})$ \textbf{then continue}
  \AlgExplain{Derive the evolution evidence for the current round from recent traces.}
  \State $\mathcal E\gets\Phi(\widetilde{\mathcal D}^{\mathrm{on}})$
  \AlgExplain{Use the large language model to form a valid registered Challenger Bundle; registration atomically assigns its stable index (online $j=1$) before evaluation or history insertion.}
  \State $\mathcal B'=\langle\pi',\chi',\ell'\rangle\gets G^{\mathrm{on}}_{\mathrm{LLM}}(\mathcal B,\mathcal E,\mathcal R^{\mathrm{on}})$
  \AlgExplain{Wait for the next feedback batch if no testable candidate is formed.}
  \State \textbf{if} $\mathcal B'=\varnothing$ \textbf{then continue}
  \AlgExplain{Perform paired evaluation of the Bundles under shared conditions in $\chi\cap\chi'$, and let $\omega$ denote the release-gate decision.}
  \State $\mathcal P\gets\Call{PairEval}{\mathcal B,\mathcal B'},\quad \omega\gets\Call{Gate}{\mathcal P,\Theta}$
  \AlgExplain{Record the candidate, evaluation result, and release decision for the current round.}
  \State $\mathcal R^{\mathrm{on}}\gets\mathcal R^{\mathrm{on}}\cup\{(\mathcal B',\mathcal P,\omega)\}$
  \AlgExplain{Mark a gate-passing candidate as the released successor asset and activate it as the current Champion only at a subsequent task boundary.}
  \State \textbf{if} $\omega=\mathrm{accept}$ \textbf{then} $\mathcal B\gets\mathcal B'$
\EndWhile
\AlgExplain{Return the policy asset and evaluation history retained after online updates.}
\State \Return $\mathcal B,\mathcal R^{\mathrm{on}}$
\end{algorithmic}
\end{olealgorithm}
\subsection{Offline Loop Policy Evolution}

Offline evolution takes archived traces and the current Champion as input, performs a multi-candidate, multi-generation evaluation--selection--generation search outside the task-execution path, and releases a candidate Bundle that passes the release gate as the successor Champion.

Archived traces $\mathcal D^{\mathrm{off}}$ yield evolution evidence $\mathcal E$ through Eq.~\eqref{eq:evolution-modes}. Based on this evidence, the large language model autonomously proposes candidates around the Champion and updates the search direction according to evaluations from previous generations. To preserve the policy identity, applicability conditions, and version provenance defined in Eq.~\eqref{eq:bundle} throughout the search, candidates are not stored as bare policies. Let $K$ be the total number of evolution generations, $\mathcal C_k$ the set of candidate Bundles at generation $k$, $\mathcal F_k$ the retained set comprising the fixed Champion and the selected nondominated candidates through generation $k$, and $\overline{\mathcal P}_k$ the cumulative paired-evaluation mapping through that generation. The offline search can then be written as
\begin{equation}
\begin{aligned}
\mathcal F_0 &= \{\mathcal B\},\\
\overline{\mathcal P}_0&=\varnothing,\\
\mathcal C_1 &=G^{\mathrm{off}}_{\mathrm{LLM}}(\mathcal B,\mathcal E,\mathcal R^{\mathrm{off}}_0),\\
\mathcal P_k&=\{\mathcal B_{k,j}\mapsto\mathcal P_{k,j}\}_{\mathcal B_{k,j}\in\mathcal C_k},\\
\mathcal P_{k,j}&=\operatorname{PairEval}(\mathcal B,\mathcal B_{k,j}),\\
\overline{\mathcal P}_k&=\overline{\mathcal P}_{k-1}\cup\mathcal P_k,\\
\mathcal A_k&=(\mathcal F_{k-1}\setminus\{\mathcal B\})\cup\mathcal C_k,\\
\mathcal F_k &=\{\mathcal B\}\cup\operatorname{Select}(\mathcal A_k,\overline{\mathcal P}_k),\\
\mathcal C_{k+1} &=V^{\mathrm{off}}_{\mathrm{LLM}}(\mathcal F_k,\mathcal E,\mathcal R^{\mathrm{off}}_k).
\end{aligned}
\label{eq:candidate-generation}
\end{equation}
where $\mathcal B=\langle\pi_c,\chi_c,\ell_c\rangle$ is the fixed Champion; $\mathcal P_k$ is the evaluation mapping for the current generation, keyed by candidate Bundles with each candidate evaluation $\mathcal P_{k,j}$ as its value; $\overline{\mathcal P}_k$ is the cumulative evaluation mapping obtained by merging each generation into the initially empty mapping $\overline{\mathcal P}_0$; and $\mathcal R^{\mathrm{off}}_k$ is the offline evaluation history. Each formed candidate is registered with a stable version index according to $\mathcal B_{k,j}\equiv\mathcal B_{v_{k,j}}$, so merging the mappings does not conflate Bundles from different generations. $\mathcal A_k$ collects retained candidates from previous generations and new candidates from the current generation but excludes the fixed Champion. Consequently, $\operatorname{Select}$ queries only candidates in the domain of the evaluation mapping, after which their union with $\mathcal B$ forms $\mathcal F_k$. $V^{\mathrm{off}}_{\mathrm{LLM}}$ generates the next generation when $k<K$. A candidate $j$ produced through recombination at generation $k$ designates one direct parent version in $\ell_{k,j}$, while other donors are recorded in the evolution evidence; its applicability conditions are restricted to those of the direct parent version and the evidence coverage. This relation preserves effective search directions and their provenance across generations.

Paired evaluation provides feedback for generational updates. Each generation's candidate Bundles are compared with the Champion on shared tasks and run conditions within the intersection of their applicability conditions and $\chi_c$. The definitions of value scores, verification risk, and resource consumption continue to follow the task contract in Section III and Eq.~\eqref{eq:loop-policy-optimization}. Benefit, reliability, tail performance, and resource cost jointly determine retention. The large language model accordingly modifies control rules or recombines complementary rules from different policies, while failures exposed by eliminated candidates also enter the evaluation history. The Champion remains a stable reference, allowing successes and failures to jointly constrain the search direction of the next generation.

Nondominated retention across generations determines only the subsequent search; formal release remains subject to Eqs.~\eqref{eq:loop-policy-optimization} and~\eqref{eq:robust-selection}. After $K$ generations of evolution, let $\mathcal F_K^{\mathrm{acc}}$ denote the candidate Bundles in the final retained set that are eligible for release. $\operatorname{ValueSelect}$ selects the candidate with the highest value score aggregated by the evaluation protocol, using reliability, tail performance, and resource cost for nondominated tie-breaking only when value scores are tied, and makes it the successor Champion $\mathcal B^{+}=\mathcal B^{\star}$. If $\mathcal F_K^{\mathrm{acc}}$ is empty, the original Champion remains unchanged. Algorithm~\ref{alg:offline-evolution} gives the procedure.

\begin{olealgorithm}{Offline Loop Policy Optimization}
\label{alg:offline-evolution}
\begin{algorithmic}[1]
\Require Champion Bundle $\mathcal B=\langle\pi_c,\chi_c,\ell_c\rangle$, archived traces $\mathcal D^{\mathrm{off}}=\{\tau_i\}$ (with $y_i$ verified by $\mathcal V_{g_i}$ and its evidence source belonging to $\mathcal S_{g_i}$), initial offline evaluation history $\mathcal R^{\mathrm{off}}_0$, number of evolution generations $K$, gate configuration $\Theta$
\Ensure Successor Champion $\mathcal B^{+}$ and offline evaluation history $\mathcal R^{\mathrm{off}}_K$ after evolution
\AlgExplain{Derive evidence related to Loop Policy modifications from archived runs.}
\State $\mathcal E\gets\Phi(\mathcal D^{\mathrm{off}})$
\AlgExplain{Use the large language model to form the initial valid candidate Bundle set; each candidate is atomically assigned stable index $v_{1,j}$ before entering the evaluation mapping.}
\State $\mathcal C_1\gets G^{\mathrm{off}}_{\mathrm{LLM}}(\mathcal B,\mathcal E,\mathcal R^{\mathrm{off}}_0)$
\AlgExplain{Initialize the elite set and cumulative evaluations with the complete Champion Bundle.}
\State $\mathcal F_0\gets\{\mathcal B\},\quad\overline{\mathcal P}_0\gets\varnothing$
\AlgExplain{Advance the multi-generation search through paired evaluation and evidence-guided policy generation.}
\For{$k=1,\ldots,K$}
  \AlgExplain{Perform Bundle paired evaluation for each candidate under applicability conditions shared with $\chi_c$, and construct the evaluation mapping for the current generation.}
  \State $\mathcal P_k\gets\varnothing$
  \ForAll{$\mathcal B_{k,j}\in\mathcal C_k$}
    \State $\mathcal P_k(\mathcal B_{k,j})\gets\Call{PairEval}{\mathcal B,\mathcal B_{k,j}}$
  \EndFor
  \AlgExplain{Accumulate the evaluations for the current generation and retain competitive policy directions from existing elite Bundles and new candidates.}
  \State $\mathcal R^{\mathrm{off}}_k\gets\mathcal R^{\mathrm{off}}_{k-1}\cup\{(\mathcal C_k,\mathcal P_k)\},\quad\overline{\mathcal P}_k\gets\overline{\mathcal P}_{k-1}\cup\mathcal P_k$
  \State $\mathcal A_k\gets(\mathcal F_{k-1}\setminus\{\mathcal B\})\cup\mathcal C_k$
  \State $\mathcal F_k\gets\{\mathcal B\}\cup\Call{Select}{\mathcal A_k,\overline{\mathcal P}_k}$
  \If{$k<K$}
    \AlgExplain{Use the large language model to mutate and recombine policies from retained Bundles and cumulative feedback; each newly formed valid Bundle is assigned stable index $v_{k+1,j}$ before next-generation evaluation.}
    \State $\mathcal C_{k+1}\gets V^{\mathrm{off}}_{\mathrm{LLM}}(\mathcal F_k,\mathcal E,\mathcal R^{\mathrm{off}}_k)$
  \EndIf
\EndFor
\AlgExplain{Collect candidates from the final retained Bundles that pass the robust release gate.}
\State $\mathcal F_K^{\mathrm{acc}}\gets\{\mathcal B'\in\mathcal F_K\setminus\{\mathcal B\}\mid\Call{Gate}{\overline{\mathcal P}_K(\mathcal B'),\Theta}=\mathrm{accept}\}$
\AlgExplain{Retain the original Champion if no candidate is eligible for release.}
\State \textbf{if} $\mathcal F_K^{\mathrm{acc}}=\varnothing$ \textbf{then} $\mathcal B^{+}\gets\mathcal B$ \textbf{and return} $\mathcal B^{+},\mathcal R^{\mathrm{off}}_K$
\AlgExplain{Select the qualified Bundle with the highest aggregate value and release it in full as the successor version of the Champion.}
\State $\mathcal B^{\star}\gets\Call{ValueSelect}{\mathcal F_K^{\mathrm{acc}},\overline{\mathcal P}_K},\quad \mathcal B^{+}\gets\mathcal B^{\star}$
\State \Return $\mathcal B^{+},\mathcal R^{\mathrm{off}}_K$
\end{algorithmic}
\end{olealgorithm}

\begin{figure*}[!t]
\centering
\includegraphics[width=0.90\textwidth]{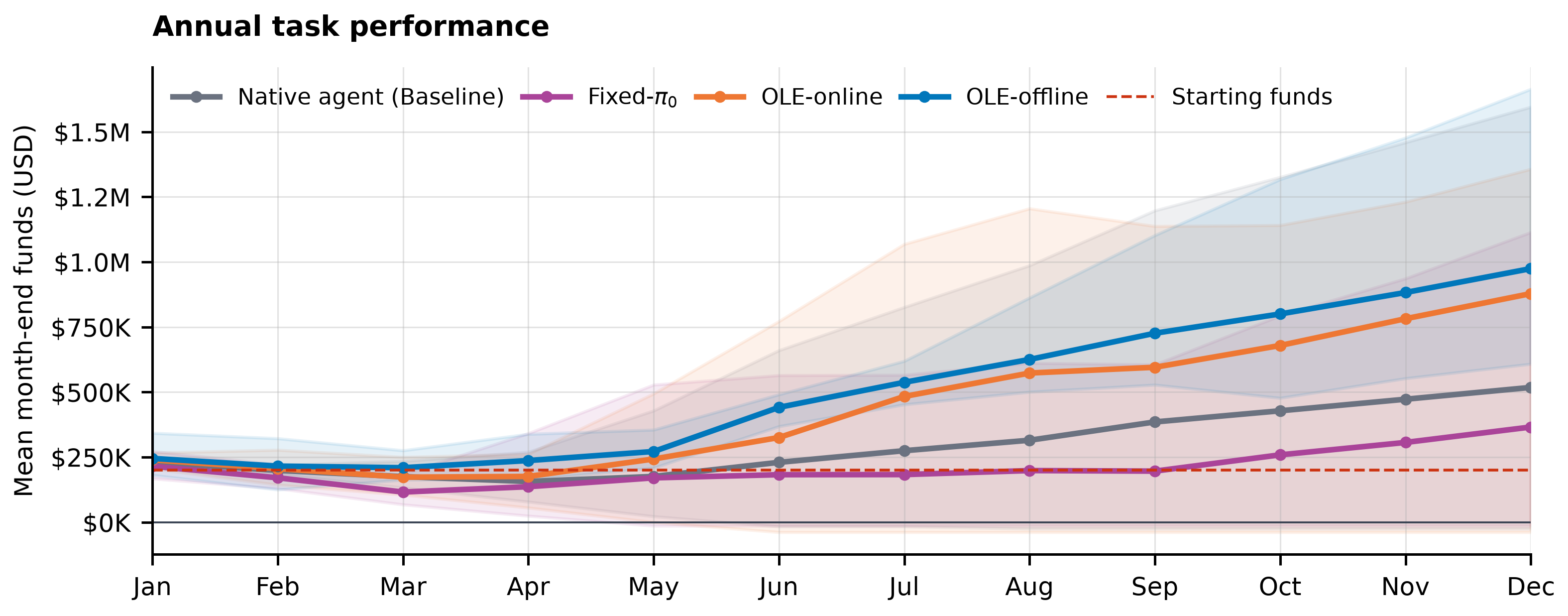}
\caption{Annual task performance of the four settings on YC-Bench. Curves and shaded regions show mean month-end funds and cross-seed ranges over three seeds, respectively; after early termination, terminal funds are carried forward only to retain a fixed aggregation denominator, not to indicate continued execution.}
\label{fig:ycbench-process}
\end{figure*}

\begin{table*}[!t]
\centering
\caption{Overall results on YC-Bench.}
\label{tab:ycbench-main-results}
\resizebox{\textwidth}{!}{%
\begin{tabular}{lcccccc}
\toprule
Method & Mean final funds (\$) & $\Delta_{\mathrm{Native}}$ & $\Delta_{\pi_0}$ & Task success rate & Annual survival & Mean maximum drawdown (\$) \\
\midrule
Native agent (Baseline) & 518,158.46 & -- & -- & 77.23\% & 1/3 & 191,462.39 \\
Fixed initial policy (Fixed-$\pi_0$) & 365,976.30 & $-29.37\%$ & -- & 73.89\% & 1/3 & 187,022.86 \\
OLE-online & 878,601.79 & +69.56\% & +140.07\% & 87.87\% & 2/3 & 166,878.69 \\
OLE-offline (fixed policy) & \textbf{974,627.52} & \textbf{+88.09\%} & \textbf{+166.31\%} & \textbf{91.80\%} & \textbf{3/3} & \textbf{115,402.09} \\
\bottomrule
\end{tabular}
}
\par\vspace*{4pt}%
\begin{minipage}{\textwidth}
\footnotesize
$\Delta_{\mathrm{Native}}$ and $\Delta_{\pi_0}$ use the Native agent and the fixed initial policy as their respective references.
\end{minipage}
\end{table*}

\section{Experimental Results}

\subsection{YC-Bench Results}

We evaluate the effect of OpenLoopEvolve on long-horizon task execution using YC-Bench~\cite{he2026ycbench}. The benchmark requires an agent to operate a business continuously over a simulated year and uses final funds to measure the cumulative effects of sequential decisions, including task selection, resource allocation, and risk management. We designate as the Baseline the Native agent that uses the official YC-Bench Agent Loop but neither integrates OpenLoopEvolve nor loads an additional Loop Policy, thereby measuring model capability under the benchmark's native execution framework. To separate the effect of introducing a policy from that of evolving it, Fixed-$\pi_0$ loads the same initial Loop Policy as the two OLE modes but keeps it unchanged throughout the year. OLE-online starts from $\pi_0$ and evolves and updates the policy online during execution, making 12 candidate-update attempts in total; OLE-offline uses a fixed policy obtained through offline evolution before the experiment and does not update it during evaluation. At each trigger, offline evolution uses the large language model to autonomously generate k=3 candidates, with a pre-registered maximum of 4 candidate rounds. All four settings use \texttt{deepseek-v4-flash}~\cite{deepseekai2026deepseekv4} under the official \texttt{medium} configuration, seeds 1/2/3, and a 20-turn context, and begin the annual evaluation from the same initial state.

Table~\ref{tab:ycbench-main-results} summarizes the annual results for the four settings. Relative to Fixed-$\pi_0$, OLE-online and OLE-offline increase mean final funds by 140.07\% and 166.31\%, respectively, and increase task success rate by 13.98 and 17.92 percentage points. Annual survival improves from 1/3 to 2/3 and 3/3, while mean maximum drawdown decreases by 10.77\% and 38.30\%, respectively. This comparison shows that both online and offline evolution can produce a more effective Loop Policy from the initial Loop Policy, with consistent advantages in long-term returns, task success rate, annual survival, and risk metrics. Fig.~\ref{fig:ycbench-process} shows the fund trajectories of the four settings over the simulated year.

\begin{table}[!t]
\centering
\caption{Main-task token usage and evolution-validation cost of OLE.}
\label{tab:ycbench-token-usage}
\scriptsize
\setlength{\tabcolsep}{2.0pt}
\begin{tabular*}{\columnwidth}{@{\extracolsep{\fill}}lrrrrr@{}}
\toprule
\multicolumn{6}{c}{(a) Main-task execution} \\
\cmidrule(lr){1-6}
Method & \shortstack{Total\\(M tokens)} & \shortstack{Mean survival\\duration (days)} & \shortstack{Average per call\\(tokens)} & \shortstack{$\Delta_{\mathrm{Base}}$\\(\%)} & \shortstack{$\Delta_{\mathrm{Fixed}}$\\(\%)} \\
\midrule
Baseline & 27.70 & 243.00 & 25,001 & -- & -- \\
Fixed-$\pi_0$ & 18.77 & 207.78 & 23,496 & $-6.02\%$ & -- \\
OLE-online & 34.70 & 294.00 & 22,029 & $-11.89\%$ & $-6.24\%$ \\
OLE-offline & 35.61 & 365.00 & 18,676 & $-25.30\%$ & $-20.51\%$ \\
\bottomrule
\end{tabular*}
\vspace{0.5em}

\begin{tabular*}{\columnwidth}{@{\extracolsep{\fill}}lrrr@{}}
\toprule
\multicolumn{4}{c}{(b) Main-task and evolution costs} \\
\cmidrule(lr){1-4}
Method & \shortstack{Main task\\(M tokens)} & \shortstack{Evolution\\(M tokens)} & \shortstack{Total\\(M tokens)} \\
\midrule
Baseline & 27.70 & 0.00 & 27.70 \\
Fixed-$\pi_0$ & 18.77 & 0.00 & 18.77 \\
OLE-online & 34.70 & 29.82 & 64.52 \\
OLE-offline & 35.61 & 24.02 & 59.63 \\
\bottomrule
\end{tabular*}
\par\vspace*{4pt}%
\begin{minipage}{\columnwidth}
\footnotesize
(a) ``Total'' is the sum of main-task tokens across the three official seeds, in millions; ``Mean survival duration'' is reported in days; and ``Average per call'' is the main-task token usage divided by the number of model calls. $\Delta_{\mathrm{Base}}$ and $\Delta_{\mathrm{Fixed}}$ denote changes in this average relative to Baseline and Fixed-$\pi_0$, respectively.

(b) Main-task totals, candidate--Champion evolution-validation costs, and their sums for the four settings, all in millions of tokens. Baseline and Fixed-$\pi_0$ perform no evolution and therefore incur zero evolution cost; the OLE totals exclude candidate-generation calls whose usage was not retained.
\end{minipage}
\end{table}

\subsection{Token Usage and Evolution Cost}

To measure changes in inference resources after introducing OLE, we aggregate the input and output tokens recorded for each model call and use their sum as token usage. All four settings include the official seeds 1/2/3. Baseline and Fixed-$\pi_0$ generate only main-task execution calls, whereas OLE-online and OLE-offline also incur candidate--Champion evolution-validation calls through paired evaluation. Table~\ref{tab:ycbench-token-usage}(a) reports main-task tokens, mean survival duration, and mean token usage per call, and computes the changes in the average relative to Baseline and Fixed-$\pi_0$. Table~\ref{tab:ycbench-token-usage}(b) compares the main-task total, evolution-validation cost, and observable total across the four settings.

Main-task execution for Baseline, Fixed-$\pi_0$, OLE-online, and OLE-offline comprises 1108, 799, 1575, and 1907 model calls, respectively, with mean survival durations of 243.00, 207.78, 294.00, and 365.00 days. The corresponding mean usage per call is 25001, 23496, 22029, and 18676 tokens. Relative to Baseline, mean usage decreases by 11.89\% and 25.30\% for OLE-online and OLE-offline, respectively; relative to Fixed-$\pi_0$, it decreases by 6.24\% and 20.51\%, respectively. Although the two OLE modes have higher main-task token totals, they also support longer execution horizons and more task calls while using fewer tokens per call on average. This indicates that, in this experiment, the Loop Policy expands the scale of sustained task execution while keeping the inference cost per interaction under control.

The online mode attempts 12 candidates, with an evolution-validation cost of 29.82M tokens and an observable total of 64.52M tokens after adding main-task usage. The offline mode has an evolution-validation cost of 24.02M tokens and an observable total of 59.63M tokens. Baseline and Fixed-$\pi_0$ perform no evolution, so their observable totals equal their main-task usage of 27.70M and 18.77M tokens, respectively. In view of the improvements in sustained execution and overall task performance, the additional evolution cost is acceptable at the scale of the current experiment. Future work will further examine whether Loop Policies remain effective across different tasks and operating environments.

\section{Conclusion}

This paper presents OpenLoopEvolve, which represents an agent's execution loop as an externalized Loop Policy that is versioned, traceable, and reusable. The framework constructs evolution evidence from run traces, uses a large language model to generate candidates, and governs policy updates through Champion--Challenger paired evaluation and robust gating. Its online and offline modes draw on recent feedback and archived experience, respectively, allowing loop improvements to accumulate beyond a single context.

Experiments on YC-Bench show that both evolution modes improve long-horizon task performance over a fixed initial policy while supporting longer effective execution with controlled inference cost per interaction. These results point to a broader role for Loop Policies as independent assets: they can be refined continually from run evidence and may be transferred and reused across tasks, agent hosts, and operating environments. Future work will evaluate this cross-scenario validity, evolution stability, and cost control at a larger scale.

\bibliographystyle{IEEEtran}
\bibliography{references}

\end{document}